\documentclass[letterpaper, 10 pt, conference]{ieeeconf}  

\IEEEoverridecommandlockouts                              

\usepackage{epsfig} 
\usepackage{amsmath} 
\usepackage{amssymb}  
\usepackage{url}
\usepackage{booktabs} 
\usepackage{cite}
\usepackage{subcaption}
\usepackage{caption}
\usepackage{graphicx}
\usepackage{xcolor}         
\usepackage{comment}
\usepackage{balance}

\newcommand{\cut}[1]{}

\title{\LARGE \bf Cross-Tool and Cross-Behavior Perceptual Knowledge Transfer \\ for Grounded Object Recognition}

\author{
    \authorblockN{Gyan Tatiya$^{1}$ \quad Jonathan Francis$^{2}$  \quad Jivko Sinapov$^{1}$}
    \thanks{$^{1}$ Department of Computer Science, Tufts University, Email: {\tt\small \{Gyan.Tatiya, Jivko.Sinapov\}@tufts.edu}.
    $^{2}$ Bosch Center for AI, Email:
    {\tt\small Jon.Francis@us.bosch.com}}
    \vspace{-1.8cm}
}

\begin{document}

\maketitle
\thispagestyle{empty}
\pagestyle{empty}

\begin{abstract}

Humans learn about objects via interaction and using multiple perceptions, such as vision, sound, and touch.
While vision can provide information about an object's appearance, non-visual sensors, such as audio and haptics, can provide information about its intrinsic properties, such as weight, temperature, hardness, and the object's sound.
Using tools to interact with objects can reveal additional object properties that are otherwise hidden (e.g., knives and spoons can be used to examine the properties of food, including its texture and consistency).
Robots can use tools to interact with objects and gather information about their implicit properties via non-visual sensors.
However, a robot's model for recognizing objects using a tool-mediated behavior does not generalize to a new tool or behavior due to differing observed data distributions.
To address this challenge, we propose a framework to enable robots to transfer implicit knowledge about granular objects across different tools and behaviors.
The proposed approach learns a shared latent space from multiple robots' contexts produced by respective sensory data while interacting with objects using tools.
We collected a dataset using a UR5 robot that performed 4,500 interactions using 6 tools and 5 behaviors on 15 granular objects and tested our method on cross-tool and cross-behavioral transfer tasks.
Our results show the less experienced target robot can benefit from the experience gained from the source robot and perform recognition on a set of novel objects.
We have released the code, datasets, and additional results: \footnotesize\url{https://github.com/gtatiya/Tool-Knowledge-Transfer}.

\end{abstract}

\section{Introduction}

Humans employ specialized tools to acquire knowledge about objects' properties and develop a comprehensive understanding of their physical characteristics, such as size, shape, texture, weight, and durability.
For example, kitchen utensils (e.g., knives, spoons) can be employed to examine the properties of food, including its texture and consistency.
Robots are expected to operate effectively in human environments; thus, the ability to estimate the physical properties of objects has become an essential component of robotics research.
Recent studies demonstrated robots can effectively use tools to interact with objects and learn about various properties, including material composition, shape, hardness, elasticity, brittleness, and adhesiveness \cite{sawhney_playing_2021, gemici_learning_2014, lenz2015deepmpc, han2020vision, bhattacharjee2019towards, gallenberger2019transfer, taunyazov2021extended, kim2023robotic, zhang2023multimodal, liu2023learning}.

Robots can use tools to execute actions on objects and observe their effects via various sensors, including visual, audio, and haptics, to acquire knowledge of objects' properties.
Non-visual modalities, such as audio and haptics, are essential, as vision alone cannot provide information about an object's intrinsic properties, including its weight, temperature, or hardness.
One of the challenges when representing non-visual modalities is that data collection requires significant time for this interactive object exploration, which may delay downstream tasks \cite{malinovska2022connectionist, li2020review, pastor2020bayesian, liu2021learning, wang2022audio, wei2021multimodal, liu2022texture}.
A logical solution for efficient learning would be to transfer object property representation to a new robot.
However, if the new robot possesses different interaction capabilities, such as new behaviors or tools, the implicit knowledge obtained by the previous robot cannot be directly transferred to the new one \cite{francis2022core}.
A robot's multisensory model for interactive perception tasks is unique to its sensors, behaviors, and tools.
Therefore, transferring knowledge of non-visual object properties across different sensorimotor contexts is challenging, and each robot must learn its task-specific sensory models from scratch.

To overcome this challenge of transferring implicit knowledge of non-visual object properties, we propose a framework leveraging triplet loss as our primary method to share tool-mediated behavioral knowledge across sensorimotor contexts, i.e., a tool-behavior pair.
Our method aims to learn a shared latent feature space by utilizing the implicit knowledge of the source robot with more experience and transferring it to the target robot with less experience.
The target robot can use the learned feature space to learn to recognize novel objects it has not previously interacted with, given the source robot has explored them.
To evaluate our method, we collected a dataset using a UR5 robot that used 6 tools to perform 5 behaviors on 15 granular objects.
We tested our method on two tasks: cross-tool transfer and cross-behavioral transfer.
Our results demonstrate the less-experienced target robot can bootstrap its object property learning by leveraging the source robot's experience.
Our method enables the target robot to recognize novel granular objects it has not interacted with before test time, thus improving its learning process' efficiency and accuracy.

\section{Related Work}

Psychological research demonstrated that children begin to comprehend how objects can be used as tools to develop intuition about the physical world at an early age \cite{bjorklund2011object}, often using utensils like spoons and forks to investigate food characteristics, such as texture.
Employing tools indicates intelligent adaptability: it necessitates an understanding of the properties of the tool and the object being acted upon \cite{connolly1989emergence}.
By using tools to explore objects, infants can modify the properties of the object being acted upon, enabling them to learn by observing the effects of their actions \cite{lockman2000perception}.

Robotics research demonstrated robots can likewise use tools to explore objects and learn about their physical properties.
Gemici {\it et al.} \cite{gemici_learning_2014} developed a method to manipulate deformable food items (e.g., bread, tofu) using kitchen tools (e.g., knife, spatula), and infer their physical properties (e.g., elasticity, adhesiveness, and hardness); their PR2 robot executed cutting and splitting actions on food items and used haptic data (e.g., force and tactile) to learn about food properties by monitoring changes in the food due to actions.
Sawhney {\it et al.} \cite{sawhney_playing_2021} used multimodal data (e.g., audio, force) to classify food materials by interacting with them using tools.
Sundaresan {\it et al.} \cite{sundaresan2022learning} deployed a multimodal policy on a Franka robot that leveraged visual and haptic observations during interaction with deformable food items to plan skewering motions rapidly and reactively.
One challenge faced by these approaches is that implicit knowledge gained by a robot via object interaction cannot be directly used by another robot, as each robot's unique sensorimotor context significantly affects the sensed data distribution and the resultant model that each robot learns.
These works focused on learning from scratch, for each robot's sensorimotor context, which is expensive at scale for robots operating under heterogeneous contexts.
We propose a framework for transferring implicit knowledge acquired during object exploration using tools, from a source robot to a target robot, which \textit{differ} in their sensorimotor contexts.

Recent studies transfer implicit knowledge across sensorimotor contexts in interactive object perception, yet they did not use tools and were limited to rigid objects \cite{tatiya2019sensorimotor, tatiya2020haptic, tatiya2023transferring}.
In \cite{tatiya2019sensorimotor}, an encoder-decoder network was used to generate a ``target'' robot's features from a ``source'' robot's learned representation for object categorization. This study only considered exploring rigid objects without tools, however, hence challenges associated with granular objects explored with tools remained unaddressed.
In \cite{tatiya2023transferring}, a distribution alignment-based approach was used to project features from two heterogeneous robots with different embodiments into a shared latent space for non-visual object property recognition. Whereas they demonstrated a shared latent space to be more effective for transfer, compared to learning projection functions to generate target context features, this study was also limited to exploring rigid objects without tools.
Moreover, they assumed heterogeneous robots had access to the same behavior in their sensorimotor context and thus learned the shared latent space for the same behavior across different robots.
To overcome these limitations, we collected a multisensory dataset using a UR5 robotic arm that performed 4,500 interactions to explore 15 granular food-like materials using 6 tools and 5 behaviors, developed a projection method for implicit knowledge transfer across two heterogeneous sensorimotor contexts, and evaluated our approach on cross-tool and cross-behavioral transfer tasks.

\section{Learning Methodology}
\label{sec:approach}

\subsection{Notation and Problem Formulation}

Consider two robots, source and target, that explore a set of granular food-like objects $\mathcal{O}$ (e.g., {\it salt}, {\it wheat}), kept in containers, by using a set of tools $\mathcal{T}$ (e.g., {\it spoon}, {\it fork}) and performing a set of exploratory behaviors $\mathcal{B}$ (e.g., {\it stirring}, {\it twist}), while recording a set of non-visual sensory modalities $\mathcal{M}$ (e.g., {\it audio}, {\it effort}).
Let the robots use each tool to perform each behavior $n$ times on each object.
Let $\mathcal{C}$ be the set of exploratory contexts, including each possible combination of a tool in $\mathcal{T}$, a behavior in $\mathcal{B}$, and a sensory modality in $\mathcal{M}$, e.g., {\it spoon-stirring-audio}, {\it fork-twist-effort}.
For the $i^{th}$ exploratory trial, the robot's observation feature is $x^{c}_{i} \in \mathbb{R}^{D_{c}}$, where $i \in \{1, ..., n\}$, $c \in \mathcal{C}$, and $D_{c}$ is the dimension of the robot's feature space under context $c$.

Let $c_s, c_t \in \mathcal{C}$ be the sensorimotor contexts of the source and target robots, respectively, which differ either by tool or behavior, e.g., for different tools, {\it spoon-stirring} as $c_s$ and {\it fork-stirring} as $c_t$, and for different behaviors, {\it spoon-stirring} as $c_s$ and {\it spoon-twist} as $c_t$; the sensory modality remains the same for both $c_s$ and $c_t$ contexts.
Consider the case where the source robot explored all objects in $\mathcal{O}$ under context $c_s$; however the target robot under context $c_t$ only explored a subset of the objects $\mathcal{O}_{shared} \subset \mathcal{O}$, and needs to learn an object recognition model for the remaining set of novel objects $\mathcal{O}_{novel} \subset \mathcal{O}$, with $\mathcal{O}_{shared} \bigcap \mathcal{O}_{novel} = \emptyset$.
Our goal is to learn a projection function using $\mathcal{O}_{shared}$, to transfer knowledge about novel objects $\mathcal{O}_{novel}$ from the more-experienced source robot to the less-experienced target robot. This knowledge transfer will help the target robot to learn about novel objects without prior interaction with them.

For transferring object knowledge, we consider a projection function $F_{{c} \rightarrow \mathcal{Z}}$, that projects the observation features from source and target contexts' feature spaces to a shared latent feature space, such that the robots can be trained to recognize objects in that latent space, as opposed to each robot's own feature space.
More specifically, $F_{c_s \rightarrow \mathcal{Z}}:x_i^{c_s} \rightarrow z_i^{c_{\mathcal{Z}}}$ and $F_{c_t \rightarrow \mathcal{Z}}:x_i^{c_t} \rightarrow z_i^{c_{\mathcal{Z}}}$, where $z_i^{c_{\mathcal{Z}}} \in \mathbb{R}^{D_{\mathcal{Z}}}$ and represents the shared latent features of size $D_{\mathcal{Z}}$.
This will enable the robots to use the observation features collected under both contexts to learn an object recognition model and perform better than a model trained only using a specific context's observation features.
Learning a shared latent feature space would enable the target robot to recognize novel objects, given the source robot has explored those objects.

\begin{figure*}[htbp]
  \centering
  \includegraphics[width=0.95\textwidth]{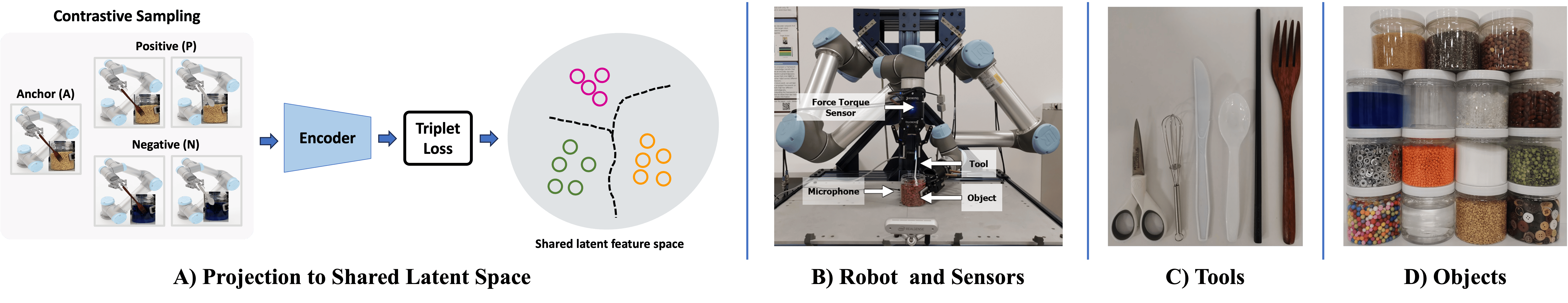}
  \vspace{-0.3cm}
  \caption{\small (A) Projection from source and target feature spaces into a shared latent space using Triplet Loss. (B) Experimental platform and sensors of the {\it UR5} robot. (C) The 6 tools used in this study: {\it metal-scissor}, {\it metal-whisk}, {\it plastic-knife}, {\it plastic-spoon}, {\it wooden-chopstick}, and {\it wooden-fork} (left to right). (D) The 15 objects used in this study (row-wise, left to right): {\it cane-sugar}, {\it chia-seed}, {\it chickpea}, {\it detergent}, {\it empty}, {\it glass-bead}, {\it kidney-bean}, {\it metal-nut-bolt}, {\it plastic-bead}, {\it salt}, {\it split-green-pea}, {\it styrofoam-bead}, {\it water}, {\it wheat}, and {\it wooden-button}.}
  \label{fig:tl_robot_tools_objects}
  \vspace{-0.5cm}
\end{figure*}

\subsection{Knowledge Transfer Model}

To learn the projection function $F_{{c} \rightarrow \mathcal{Z}}$, we employ Triplet Loss (TL) \cite{balntas2016learning}, which guides our neural projection to map sensory data from both source and target contexts ($c_s, c_t$) into a common latent space (Fig. \ref{fig:tl_robot_tools_objects}A).
The essence of triplet loss is to ensure that embeddings of examples belonging to the same object class are closer in the latent space than those of dissimilar examples from different object classes:%
\begin{equation}
\begin{split}
    & \mathcal{L}(A, P, N) = \min(0, d(A, P) - d(A, N) + \alpha),
\label{eq:triplet_loss}
\end{split}
\end{equation}%
\noindent for anchor example $A$, positive example $P$, negative example $N$, and a margin hyperparameter $\alpha$ that defines a minimum difference that must be maintained between the distance from the anchor to the positive sample and the distance from the anchor to the negative sample (we set $\alpha = 1$).
The function $d(x,y)$ calculates the distance between examples $x$ and $y$ using the Euclidean distance formula, $ d(x, y) = \sqrt{\sum_{i=1}^{n} (x_i - y_i)^2} $, where $x$ and $y$ are the two examples being compared, and $n$ is the dimensionality.

To train our projection function using triplet loss, we construct a dataset of triplets ($A,P,N$) as follows:
For each object $o \in \mathcal{O}$, we designate the anchor ($A$) as data from the source context ($c_s$) for object o.
The positive ($P$) is either from the same source context ($c_s$) but a different trial or from the target context ($c_t$) for the same object.
The negative ($N$) is either from the same source context ($c_s$) or from the target context ($c_t$) for a different object.
We randomly sample a single example for both positive and negative cases when multiple examples are available from source and target contexts.
This triplet dataset is created using all trials of objects, and we optimize the triplet loss function over it.
By doing so, our network learns to map sensory data from both source and target contexts into a shared latent space ($\mathcal{Z}$).
In this latent space, objects of the same class are brought closer together than objects of different classes.
Consequently, when the target robot encounters novel objects ($\mathcal{O}_{shared}$), it can effectively recognize them by comparing their embeddings in the shared latent space ($\mathcal{Z}$), even if it has not directly interacted with them during exploration.
This process ensures that the robot builds a robust representation of objects, capable of generalizing across diverse contexts and effectively recognizing novel objects.

\subsection{Model Implementation}

The knowledge transfer model is constructed as a Multi-Layer Perceptron (MLP) comprising three hidden layers with 1000, 500, and 250 units, employing the Rectified Linear Unit (ReLU) activation function.
This model projects sensory data into a shared latent vector of dimension $D_{\mathcal{Z}}$ = 125.
To enable the target robot to recognize novel objects, we use shared latent features corresponding to the novel objects in the target context projected by the source context.
These shared latent features are comparable and can be employed to train a standard multi-class classifier across different contexts.
We train an MLP model with a single hidden layer of 100 units for the recognition task, allowing the target robot to discern objects it has not directly encountered.
The knowledge transfer and classification models are updated for 500 training epochs, leveraging the Adam optimization algorithm \cite{kingma_adam_2015} with a learning rate set at $10^{-4}$.
We used PyTorch \cite{paszke2019pytorch} for model implementation.

\section{Evaluation Design}

\subsection{Experimental Platform and Feature Extraction}

\subsubsection{Robot and Sensors}

We collected a dataset using the {\it UR5} robot with a 6-DOF and a 2-finger Robotiq 85 gripper (shown in Fig. \ref{fig:tl_robot_tools_objects}B).
The {\it UR5} had a Seeed Studio ReSpeaker microphone placed on its workstation, and a force sensor measuring effort at each joint, and a force-torque sensor at the end-effector.
We recorded audio data at a sampling rate of 16 kHz, effort data at 135 Hz, and force data at 125 Hz.

\subsubsection{Tools, Exploratory Behaviors and Objects}

\begin{figure}
\centering
\includegraphics[width=0.9\linewidth]{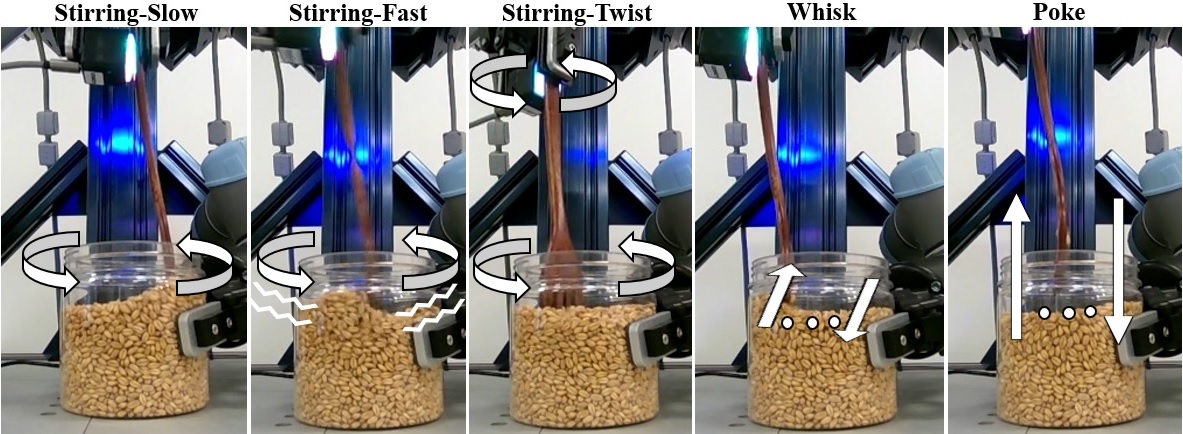}
\vspace{-0.1cm}
\caption{\small The 5 behaviors used to explore objects: {\it stirring-slow}, {\it stirring-fast}, {\it stirring-twist}, {\it whisk}, and {\it poke} (left to right).}
\label{fig:behaviors}
\vspace{-0.6cm}
\end{figure}

The robot used 6 tools {\it metal-scissor}, {\it metal-whisk}, {\it plastic-knife}, {\it plastic-spoon}, {\it wooden-chopstick}, and {\it wooden-fork} (Fig. \ref{fig:tl_robot_tools_objects}C) to perform 5 interactive behaviors: {\it stirring-slow}, {\it stirring-fast}, {\it stirring-twist}, {\it whisk}, and {\it poke} (Fig. \ref{fig:behaviors}).
We chose these specific tools and behaviors because they capture different aspects of objects' properties.
The interactive behaviors are encoded as robot joint-angle trajectories.
The robot explored 15 objects: {\it cane-sugar}, {\it chia-seed}, {\it chickpea}, {\it detergent}, {\it empty}, {\it glass-bead}, {\it kidney-bean}, {\it metal-nut-bolt}, {\it plastic-bead}, {\it salt}, {\it split-green-pea}, {\it styrofoam-bead}, {\it water}, {\it wheat}, and {\it wooden-button} (Fig. \ref{fig:tl_robot_tools_objects}D) kept in cylindrical containers.


\subsubsection{Data Collection}

While recording sensory data, the robot performed all 5 behaviors in a sequence on an object using a tool.
Once an object was explored using a tool, the same object was not explored again until all the objects were explored using that tool to limit any transient noise effects after a trial on an object.
We used another {\it UR5} arm only to hold the containers (Fig. \ref{fig:tl_robot_tools_objects}B).
The robot performed 10 trials on each object using a tool, resulting in 4,500 interactions (6 tools x 5 behaviors x 15 objects x 10 trials).
Datasets download link, source code, and complete results are available on the GitHub page of the study.

\subsubsection{Feature Extraction and Data Augmentation}

\begin{figure}
\centering
\includegraphics[width=0.95\linewidth]{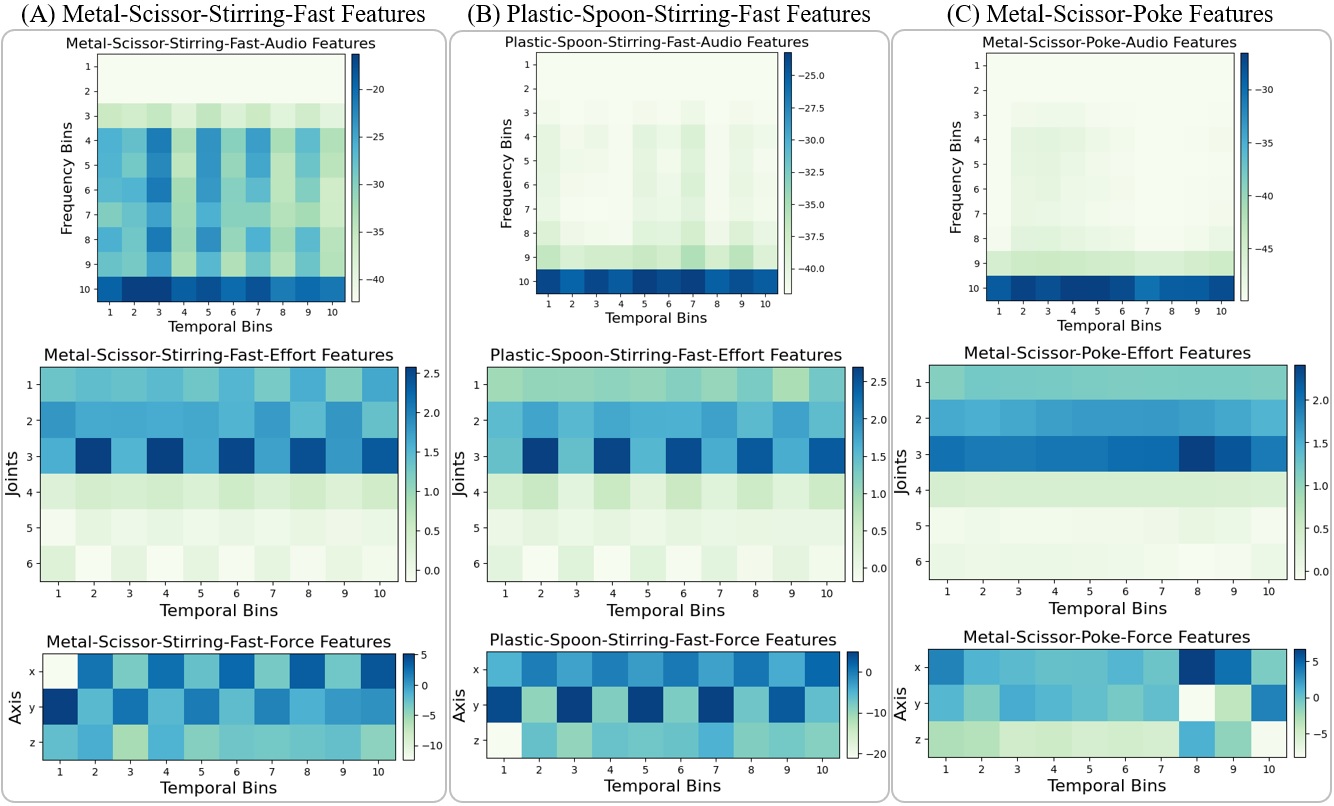}
\caption{\small Examples of {\it audio}, {\it effort}, and {\it force} features (top to bottom) when {\it UR5} uses {\it metal-scissor} tool to perform {\it stirring-fast} (A) and {\it poke} (C) behaviors, and uses {\it plastic-spoon} tool to perform {\it stirring-fast} behavior (B) on a {\it metal-nut-bolt} object. Please note the difference in the features when only the tools are different ((A) and (B)) and when only the behaviors are different ((A) and (C)).}
\label{fig:Features}
\vspace{-0.5cm}
\end{figure}

We used the 6 tools and 5 interactive behaviors listed above to conduct our experiments.
We used 3 non-visual modalities (i.e., audio, effort, and force) because they are essential for the human somatosensory perception of object properties.
The feature extraction parameters and data augmentation routines were adopted from \cite{tatiya2023transferring}. 
To represent audio data, first, we used librosa \cite{mcfee_librosa_2015} to generate mel-scaled spectrograms of audio wave files recorded by robots with FFT window length 1024, hop length 512, and 60 mel-bands.
Secondly, a spectro-temporal histogram was computed by discretizing both time and frequencies into 10 equally-spaced bins, where each bin consisted of mean of values in that bin.
Similarly, we discretized time into 10 equally-spaced bins for effort and force data, for 6 joints and 3 axes, respectively.
Thus, audio, effort, and force data are represented as 100, 60, and 30 dimensional feature vectors, respectively.
Fig. \ref{fig:Features} visualizes the robot's {\it audio}, {\it effort}, and {\it force} features when it uses different tools to perform different behaviors on an object.
To augment data, we computed the mean and standard deviation of each bin in the discretized representation of all trials of an object, and sampled $10$ additional trials of each object.
These augmented data were used to train all methods.

\subsection{Evaluation}

\subsubsection{Transfer and Baseline Conditions}

For the transfer condition, we assume the source robot interacts with all 15 objects in $\mathcal{O}$ under the source context, but the target robot interacts with only 10 randomly selected objects ($66.67\%$ of objects) in $\mathcal{O}_{shared}$ under the target context.
The 10 shared objects under both contexts are used to train the knowledge transfer model that projects the sensory signal of both contexts into a shared latent space.
Subsequently, an object classifier is trained using the projected data from the source context corresponding to the 5 objects ($33.33\%$ of objects) in $\mathcal{O}_{novel}$ that are novel to the target context.
We used the latent features corresponding to the 5 novel objects under the target context generated by the trained knowledge transfer model to test the object classifier.
We used two baseline conditions.
For baseline 1, the target robot is trained to recognize objects using its own data collected during object interactions under the target context.
This baseline would show the target robot's performance if it actually explored all the objects under the target context during the training phase.
Baseline 2 is similar to baseline 1, except the target robot is trained under the source context.
The target robot's own data observed under the target context are used to test both baseline conditions.
Baseline 2 is zero-shot classification, as the target robot is trained under the source context and tested under the target context.
In each condition, the classifiers were trained on randomly sampled 8 trials ($80\%$ of trials) from each of the 5 novel objects and tested on the held-out 2 trials ($20\%$ of trials).
The process of randomly selecting 10 objects in $\mathcal{O}_{shared}$ to train knowledge transfer mode, training and testing the object classifiers on 5 novel objects for the transfer and baseline conditions, was repeated 10 times to compute performance statistics.
We used a dataset with one robot; hence, we assumed the source and target robots are physically identical, although employing different tools and behaviors during object interaction.
Nonetheless, our proposed transfer learning methodology remains pertinent in scenarios where the two robots are not physically identical.

\subsubsection{Evaluation Metrics}

We used two metrics to evaluate the object recognition performance of the target robot on the objects it has not explored.
First is accuracy, defined as $A = \frac{\text{correct \; predictions}}{\text{total predictions}}$ (\%).
The second metric is accuracy delta ($A\Delta$), which measures the difference in classification accuracy by using the latent features for training instead of the ground-truth features. 
We define accuracy delta as $A\Delta = A_{truth} - A_{latent}$, where $A_{truth}$ and $A_{latent}$ are the accuracies obtained when using ground-truth and latent features, respectively.
A smaller accuracy delta indicates it is easy for the target robot to learn about the novel objects using the knowledge transferred by the source robot.
To report both metrics' results, we use the recognition accuracy computed by performing a weighted combination of each modality used based on their performance on the training data.

\subsubsection{Transfer Tasks}

We consider two tasks: cross-tool sensorimotor transfer and cross-behavioral sensorimotor transfer.
In cross-tool sensorimotor transfer, the source and target robots' contexts differ only by tools (e.g., {\it scissor-stirring} as the source context and {\it spoon-stirring} as the target context).
In cross-behavioral sensorimotor transfer, the source and target robots' contexts differ only by behaviors (e.g., {\it scissor-stirring} as the source context and {\it scissor-poke} as the target context).
In both tasks, we align the same modality for both source and target robots into the shared latent space.

\subsubsection{Baseline Transfer Method}

We use Kernel Manifold Alignment (KEMA) \cite{tuia_kernel_2016} as our baseline.
KEMA is a distribution alignment method to align observation features from various contexts and represent them within a shared latent space.
KEMA constructs domain-specific projection functions, which project data from both source and target contexts into a shared latent space.
This projection ensures that examples of the same object class are closely grouped while those from different classes are separated.
KEMA has demonstrated effectiveness in various domains, including visual object recognition \cite{tuia_kernel_2016}, facial expression recognition \cite{tuia_kernel_2016}, and human action recognition \cite{liu2018transferable}.
In robotics, KEMA has been successfully employed to align haptic data \cite{tatiya2020haptic} and audio data \cite{tatiya2023transferring} across heterogeneous robots.

\section{Results}

\subsection{Illustrative Example}

\begin{figure*}[htbp]
  \centering
  \includegraphics[width=0.90\textwidth]{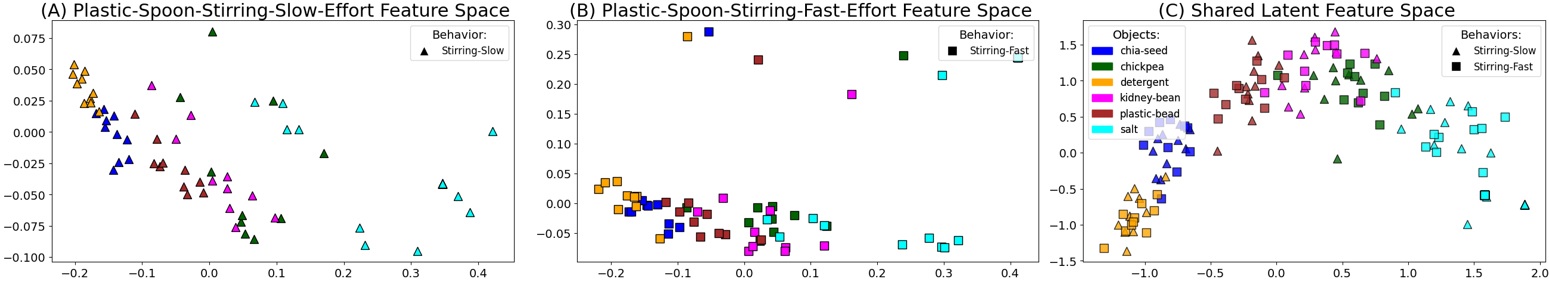}
  \caption{\small Original sensory features of (A) {\it plastic-spoon-stirring-slow} and (B) {\it plastic-spoon-stirring-fast} for {\it effort} performed on 6 objects in 2D space, and first 2 dimensions of corresponding features in the shared latent feature space (C).}
  \label{fig:illustrative_example}
  \vspace{-0.7cm}
\end{figure*}

Consider the case where {\it UR5} uses the {\it plastic-spoon} to perform the {\it stirring-slow} behavior as the source context and the {\it stirring-fast} behavior with the same tool as the target context, on 15 objects, 10 times, while recording {\it effort} signals, which our knowledge transfer model uses to generate the shared latent features.
Fig. \ref{fig:illustrative_example} visualizes the original and latent features of 6 objects in 2D space.
To visualize the original sensory signal of source and target contexts, we reduced their dimension to 2 by Principal Component Analysis (PCA) and plotted in Fig. \ref{fig:illustrative_example}A and \ref{fig:illustrative_example}B.
Similarity, we plot the features in 2D for visualization (Fig. \ref{fig:illustrative_example}C): datapoints collected in both contexts are clustered together in the shared latent space, indicating both feature spaces are aligned effectively.

Consider another case where {\it UR5} learns to recognize the 15 objects using each tool and each behavior.
To achieve this, we train a classifier for each tool and behavior pair using 8 trials of each object and test it on the held-out 2 trials.
We perform 5-fold cross-validation such that each trial of the 15 objects is included in the test set once and compute the mean accuracy of all folds.
Table \ref{tab:object_classification_results} shows the recognition accuracy computed by performing a weighted combination of all 3 non-visual modalities used based on their performance on the training data.
In the table, the bottom row shows the accuracy computed by performing a weighted combination of all the behaviors.
We report these recognition accuracies to illustrate how each tool and behavior pair performs.

\begin{table}
\caption{\small Accuracy percentage (\%) achieved by {\it UR5} using each tool and behavior pair to recognize 15 objects ($\uparrow$).}
\vspace{-0.1cm}
\label{tab:object_classification_results}
\scriptsize
\centering
\resizebox{0.95\columnwidth}{!}{
\begin{tabular}{p{0.25\linewidth}p{0.08\linewidth}p{0.08\linewidth}p{0.08\linewidth}p{0.08\linewidth}p{0.08\linewidth}p{0.08\linewidth}}
\toprule
 & \multicolumn{6}{c}{\textbf{\textsc{{\it \textbf{Tools}}}}} \\
\cmidrule(r){2-7}
{\it \textbf{Behaviors}} & {\it \textbf{Metal-Scissor}} & {\it \textbf{Metal-Whisk}} & {\it \textbf{Plastic-Knife}} & {\it \textbf{Plastic-Spoon}} & {\it \textbf{Wooden-Chopstick}} & {\it \textbf{Wooden-Fork}} \\
\midrule
Stirring-Slow & 26.00 & 33.33 & 18.67 & 36.67 & 16.00 & 42.00 \\
Stirring-Fast & 48.67 & 35.33 & 30.00 & 44.67 & 18.00 & 42.00 \\
Stirring-Twist & 17.33 & 26.00 & 16.00 & 23.33 & 15.33 & 32.67 \\
Whisk & 20.00 & 24.00 & 27.33 & 28.00 & 14.67 & 39.33 \\
Poke & 20.00 & 15.33 & 21.33 & 24.67 & 17.33 & 30.00 \\
\midrule
All behaviors & 51.33 & 50.00 & 48.00 & 52.00 & 39.33 & 63.33 \\
\bottomrule
\end{tabular}
}
\vspace{-0.7cm}
\end{table}

\subsection{Accuracy Results of Object Recognition}

For cross-tool sensorimotor transfer, each of the 6 tools is projected to all the other 5 tools, for each \textit{behavior}, allowing 150 cross-tool projections (6 tools $\times$ 5 other tools $\times$ 5 behaviors).
For cross-behavioral sensorimotor transfer, each of the 5 behaviors is projected to all the other 4 behaviors, for each \textit{tool}, allowing 120 cross-behavioral projections (5 behaviors $\times$ 4 other behaviors $\times$ 6 tools).
Table \ref{tab:accuracy_results} shows the mean accuracy and $A\Delta$ values for all projections, considering both transfer methods (TL and KEMA) and both baseline conditions.
Our transfer method (TL) achieves higher accuracy than the baseline condition 1 in 74 and 8 projections across all cross-tool and cross-behavioral projections, respectively.
In comparison to KEMA, our method achieves a lower mean $A\Delta$ in both baseline conditions.
These results show using latent features transferred by the source robot using our method aids the target robot in learning a recognition model that generalizes better for object recognition under specific projections (discussed in the next section).
Notably, a smaller mean $A\Delta$ (including negative mean $A\Delta$) in Table \ref{tab:accuracy_results} indicates it is easy for the target robot to learn a classifier from latent features projected by the source robot and achieve comparable performance as if the target robot actually explored the objects.

\begin{table}
\caption{\small Mean accuracy ($\uparrow$) and $A\Delta$  ($\downarrow$) for transfer and both baseline conditions in cross-tool and cross-behavior transfers.}
\label{tab:accuracy_results}
\scriptsize
\centering
\resizebox{0.95\columnwidth}{!}{
\begin{tabular}{p{0.32\linewidth}p{0.13\linewidth}p{0.19\linewidth}p{0.17\linewidth}p{0.19\linewidth}}
\toprule
 & \multicolumn{2}{c}{\it \textbf{KEMA}} & \multicolumn{2}{c}{\it \textbf{TL (ours)}} \\
\cmidrule(r){2-3}
\cmidrule(r){4-5}
 & {\it \textbf{Cross-Tool}} & {\it \textbf{Cross-Behavior}} & {\it \textbf{Cross-Tool}} & {\it \textbf{Cross-Behavior}} \\
\midrule
Baseline 1 Mean Accuracy & 50.6$\pm$12.5\% & 50.7$\pm$12.3\% & 50.6$\pm$12.3\% & 50.3$\pm$12.4\% \\
Baseline 2 Mean Accuracy & 26.5$\pm$6.0\% & 23.9$\pm$4.7\% & 26.0$\pm$5.8\% & 24.1$\pm$4.9\% \\
Transfer Mean Accuracy & 22.3$\pm$4.3\% & 22.1$\pm$4.9\% & 49.9$\pm$10.6\% & 33.7$\pm$8.6\% \\
\midrule
Baseline 1 Mean $A\Delta$ & 28.3$\pm$13.8\% & 28.6$\pm$14.6\% & \textbf{0.7$\pm$13.9\%} & \textbf{16.5$\pm$12.9\%} \\
Baseline 2 Mean $A\Delta$ & 4.2$\pm$8.2\% & 1.7$\pm$7.3\% & \textbf{-23.8$\pm$11.5\%} & \textbf{-9.5$\pm$7.9\%} \\
\bottomrule
\end{tabular}
}
\vspace{-0.2cm}
\end{table}

We conducted additional experiments to assess the robustness and adaptability of our method.
First, we repeated the experiments without data augmentation, to simulate 
cases with limited data availability.
In this context, when using KEMA, the mean $A\Delta$ for baseline condition 1 averaged 29.5$\pm$13.9\% and 30.6$\pm$13.5\% for all cross-tool and cross-behavioral projections, respectively.
In contrast, our method (TL) achieved substantially lower mean $A\Delta$ values of 1.9$\pm$12.0\% and 15.8$\pm$11.6\% for the same projections.
Remarkably, even without data augmentation, our method consistently outperforms KEMA for both baseline conditions.
Furthermore, we explored the impact of using a simple SVM classifier, as an alternative to an MLP, in the same experiments, both with and without data augmentation.
Regardless of the classifier used, our method consistently achieved lower mean $A\Delta$ values for baseline conditions 1 and 2 compared to KEMA. These results underscore the robustness of our approach across varying data availability scenarios and with different classification models, demonstrating its ability to learn effective latent features that significantly aid the target robot in recognizing novel objects under diverse conditions.

\subsection{Accuracy Delta Results of Object Recognition}

In cross-tool projections, for each tool as the source tool, we used all the other tools as the target tool, allowing 30 (6 tools $\times$ 5 other tools) projections for each behavior.
In cross-behavioral projections, for each behavior as the source behavior, we used all the other behaviors as the target behavior, allowing 25 (5 behaviors $\times$ 4 other behaviors) projections for each tool.
Table \ref{tab:accuracy_delta_results} shows the mean $A\Delta$ (baseline 1) of all projections for each behavior and each tool in cross-tool and cross-behavioral projections, respectively.

\begin{table}
\caption{\small Mean $A\Delta$ (baseline 1) for each behavior in cross-tool projections and for each tool in cross-behavioral projections ($\downarrow$).}
\vspace{-0.1cm}
\label{tab:accuracy_delta_results}
\scriptsize
\resizebox{\columnwidth}{!}{
\centering
\begin{tabular}{p{0.25\linewidth}p{0.25\linewidth}p{0.25\linewidth}p{0.25\linewidth}}
\toprule
\multicolumn{2}{c}{\textsc{{\it \textbf{Cross-Tool}}}} & \multicolumn{2}{c}{\textsc{{\it \textbf{Cross-Behavior}}}} \\
\textbf{Behaviors} & {\textsc{{\it \textbf{Mean $A\Delta$}}}} & \textbf{Tools} & \textbf{Mean $A\Delta$} \\
\cmidrule(r){1-2} \cmidrule(r){3-4}
Stirring-Slow & 6.6$\pm$12.0\% & Metal-Scissor & 11.1$\pm$9.5\% \\
Stirring-Fast & 12.2$\pm$14.4\% & Metal-Whisk & 14.1$\pm$11.7\% \\
Stirring-Twist & -6.6$\pm$12.8\% & Plastic-Knife & 16.2$\pm$8.2\% \\
Whisk & -7.8$\pm$11.1\% & Plastic-Spoon & 25.6$\pm$12.5\% \\
Poke & -1.4$\pm$5.3\% & Wooden-Chopstick & 5.6$\pm$9.7\% \\
 --- & --- & Wooden-Fork & 26.6$\pm$11.0\% \\
\bottomrule
\end{tabular}
}
\vspace{-0.6cm}
\end{table}


For cross-tool projections, the least mean $A\Delta$ is achieved by {\it whisk}, and {\it stirring-twist} behaviors.
Compared to other behaviors, these behaviors deform the tools less during object interaction and are shorter behaviors.
However, longer behaviors deform the tools more with object interaction and achieve higher mean $A\Delta$ (e.g., {\it poke}, {\it stirring-slow}, and {\it stirring-fast}).
This shows if the robot needs to use a new tool, the prior experience gained by a shorter behavior that deforms the tools less would be better to be transferred to the target context with the new tool.
For cross-behavioral projections, the least mean $A\Delta$ is achieved by {\it wooden-chopstick}, {\it metal-scissor}, and {\it metal-whisk} tools.
Compared to other tools, these tools get deformed less while performing behaviors on objects and have pointed ends making limited object contact.
However, other tools have wider ends, making them deform more with object interaction (e.g., {\it plastic-knife}, {\it plastic-spoon}, and {\it wooden-fork}).
This shows that if the robot needs to perform a new behavior, the prior knowledge of behaviors gained using rigid and pointed tools would be better to be transferred to the target context's new behavior.

\subsection{Tools and Behaviors Transfer Relationships}

\begin{figure}
\centering
\includegraphics[width=0.9\linewidth]{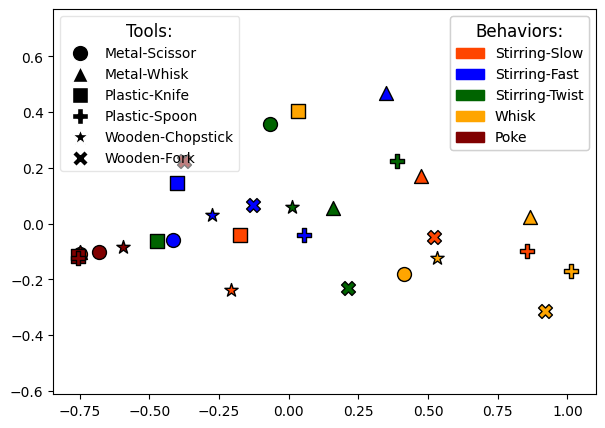}
\vspace{-0.3cm}
\caption{\small 2D PCA embedding of the $A\Delta$ matrix for cross-tool and cross-behavior projections. Every point stands for a context (i.e., a tool and behavior pair). Closer points reflect contexts across which knowledge transfer is more efficient.}
\label{fig:tools_behaviors_2D_neighborhood_graph}
\vspace{-0.7cm}
\end{figure}

To compute the transfer relation between each tool and behavior pair, we consider cross-tool and cross-behavior projections simultaneously.
For such projections, for each tool and behavior pair as the source context, we use all the other tool and behavior pairs as the target context.
More specifically, we used 30 tool and behavior pairs (6 tools $\times$ 5 behaviors) as the source context and the other 35 pairs as the target context, allowing 870 (30 $\times$ 29) projections.
We compute the $A\Delta$ for each projection and represent them in an 870 $\times$ 870 matrix, where the $A\Delta$ of identical contexts is 0.
A 2D visualization of PCA embedding of the $A\Delta$ matrix is shown in Fig. \ref{fig:tools_behaviors_2D_neighborhood_graph}.
Each dot in the plot represents a context, and the distance between a pair of contexts indicates how efficient the transfer is between them.
The closer the two contexts are, the more efficiently they transfer knowledge.

Contexts with the same or similar behaviors are clustered together, suggesting the source and target contexts with similar behaviors and different tools transfer better.
The most tightly clustered behavior is {\it poke}.
Similar behaviors are loosely clustered together (i.e., {\it stirring-fast}, {\it stirring-twist}, and {\it stirring-slow}).
Non-deformable tools (i.e., {\it metal-scissor} and {\it wooden-chopstick}) with a behavior in the source context are closer to the other behaviors in the target contexts.
This indicates that non-deformable tools capture similar object properties across different behaviors, as such tools are less impacted by different behaviors during object interaction.
These findings are consistent with the cross-tool and cross-behavior transfers' results previously outlined.

\section{Conclusion and Future Work}

Robots can acquire implicit knowledge about object properties by performing tool-mediated behaviors on granular objects and processing non-visual modalities.
However, representing implicit knowledge for each different sensorimotor context can be expensive, as it necessitates the robots to explore objects from scratch in each new context.
To overcome this challenge, we proposed a framework for transferring implicit object property knowledge across different sensorimotor contexts.
We evaluated the effectiveness of our approach on cross-tool and cross-behavioral transfer tasks.
Our results demonstrated that transferring implicit knowledge from the source robot to the target robot accelerates the target robot's learning, even if it has explored fewer objects.

Our framework encoded different behaviors in the robot for object exploration using tools, but our future work aims to enable robots to learn behaviors for object interaction autonomously.
We assumed that both source and target contexts explored objects using the same modality, and we used this modality while learning the shared latent space.
We plan to perform cross-modality projections especially for cases where the target robot has a different non-visual sensor from the source robot and select sensorimotor contexts for learning projections more efficiently.
We aim to automate the selection of objects to be explored to learn an effective projection faster.
We envision multiple source contexts transferring their knowledge to the target context.
In conclusion, our proposed framework can transfer implicit knowledge about objects from one robot's sensorimotor context to another, leading to accelerated learning in the target context.

\clearpage

\balance 

\bibliographystyle{IEEEtran}
\bibliography{references}

\end{document}